\documentclass{article}
\usepackage[utf8]{inputenc}
\usepackage{graphicx}
\usepackage[english]{babel}
\usepackage[table]{xcolor}
\usepackage{multirow}
\usepackage{adjustbox}
\usepackage{geometry}
\usepackage{amsfonts}
\usepackage{nicefrac}
\usepackage{amssymb}
\usepackage{newtxmath}
\usepackage{scrextend}
\usepackage{numprint}
\usepackage{microtype}
\usepackage{graphicx}
\usepackage[justification=centering]{caption}
\usepackage{subcaption}
\usepackage{setspace}  % For double-spacing
\usepackage[style=numeric-comp,sorting=none,maxnames=6,minnames=3,backend=biber]{biblatex}
\usepackage{booktabs}
\usepackage{multirow}
\usepackage{tabularx}
\usepackage{siunitx}
\usepackage{hyperref}
\usepackage{chngpage}
\usepackage{comment}
\usepackage{makecell}
\usepackage{nicematrix,enumitem,booktabs}
\usepackage[para]{threeparttable}
\usepackage{tikz}
\usepackage{array}
\usepackage[capitalise]{cleveref}

\graphicspath{ {images/} }

\begin{document}

\begin{titlepage}
    \centering
    \vspace*{1cm}

    {\Huge \bfseries Detecting Daily Living Gait Amid Huntington's Disease Chorea using a Foundation Deep Learning Model \par}

    \vspace{1.5cm}
    
    \textbf{Authors:} \\
    Dafna Schwartz\textsuperscript{1,2} \\
    Lori Quinn\textsuperscript{3} \\
    Nora E. Fritz\textsuperscript{4}\\
    Lisa M. Muratori\textsuperscript{5}\\
    Jeffery M. Hausdorff\textsuperscript{2,6} \\
    Ran Gilad Bachrach\textsuperscript{1} \\

    \vspace{1cm}

    \textbf{Affiliations:} \\
    \textsuperscript{1}Department of Bio-Medical Engineering, Tel Aviv University  \\
    \textsuperscript{2}Center for the Study of Movement, Cognition, and Mobility (CMCM), Neurological Institute, Tel Aviv Sourasky Medical Center, Tel Aviv, Israel \\
    \textsuperscript{3}Department of Biobehavioral Sciences, Teachers College, Columbia University, NY, USA\\
    \textsuperscript{4}Departments of Health Care Sciences and Neurology, Wayne State University, Detroit, MI, USA\\
    \textsuperscript{5}Doctor of Physical Therapy Program, Stony Brook University, Stony Brook, NY, USA\\
    \textsuperscript{6}Department of Physical Therapy, Faculty of Medical \& Health Sciences  and Sagol School of Neuroscience, Tel Aviv University; Rush Alzheimer’s Disease Center and Department of Orthopaedic Surgery, Rush University Medical Center, Chicago, IL\\

    \vspace{1.5cm}

    \vfill

\end{titlepage}

\section*{Abstract}
\textbf{Background:} Wearable sensors offer an objective, non-invasive means to collect physical activity (PA) data with walking as a key component. However, in conditions involving involuntary movements, for example, due to neurodegenerative diseases (NDDs), existing models struggle to accurately detect gait bouts. We developed a novel gait detection approach for individuals with movement disorders, like Huntington’s disease (HD), in daily-living environments.\\
\textbf{Methods:} We developed J-Net, a deep learning model inspired by U-Net, which uses a pre-trained self-supervised foundation model fine-tuned with in-lab HD data and paired with a segmentation head for sample-wise gait classification, to detect gait during daily living using wrist-worn accelerometer data. J-Net’s performance was evaluated on in-lab and daily-living data from individuals with HD, Parkinson’s disease (PD) patients, and healthy controls.\\
\textbf{Results:} J-Net outperformed existing methods across all in-lab metrics, achieving a 10-percentage point increase in ROC-AUC for people with HD, reaching 0.97. In daily-living settings, J-Net did not show significant differences in median daily walking time between HD and controls ($p = 0.23$), while other models suggested, counterintuitively, that people with HD walked significantly more ($p < 0.005$). Median walking time measured by J-Net correlated with the UHDRS-TMS clinical measure of HD severity ($r=-0.52; p=0.02$), supporting its clinical meaningfulness. Fine-tuning J-Net on PD data also improved gait detection over existing methods.\\
\textbf{Conclusions:} The J-Net architecture is a promising tool for gait detection in individuals with NDDs like HD and PD using wrist-worn accelerometers. By combining a foundation model with a segmentation network, J-Net outperformed existing methods, particularly in detecting gait during severe chorea in HD. In daily-living settings, J-Net provided an estimate of walking time and insights on the diurnal patterns of gait with face validity, in contrast to existing methods. The novel dataset created for this study, is now open source, along with the J-Net model, enabling further research.

\section{Introduction}
Physical activity (PA) is essential for maintaining health, reducing morbidity and mortality, and enhancing quality of life. Traditionally, self-report has been used to evaluate PA\cite{strain2024national}, though subjective measures often lack a strong correlation with objective measures\cite{prince2008comparison}, limiting utility. Wrist-worn devices and smartphones provide objective, non-invasive means to collect PA data\cite{migueles2017accelerometer}, particularly for walking, a key component of PA. However, an important consideration that has largely been ignored is that among people with involuntary movements, such as those with neurodegenerative diseases (NDDs), existing models struggle to detect gait bouts accurately.

NDDs like Huntington’s disease (HD) and Parkinson’s disease (PD) are characterized by progressive motor impairments\cite{walker2007huntington,moustafa2016motor}, posing unique challenges to gait identification from wearable sensors. In PD, motor symptoms include involuntary hand movements such as tremor and dyskinesia, while HD is marked by chorea and dystonia, manifesting as involuntary hyperkinetic and hypokinetic movements. Recent work has explored chorea quantification in HD using wearable sensors during structured tasks in daily-living settings\cite{gordon2020quantification}, demonstrating feasibility. However, this study does not allow continuous measurement of chorea or gait in HD. Accurate identification of walking is crucial, as gait metrics serve as sensitive indicators of PA and disease severity\cite{galperin2020sensor,gassner2020gait,hillel2019every,warmerdam2020long}. Moreover, objectively assessing gait in daily living through wrist-worn sensors provides valuable insights into intervention effectiveness. The present study focuses on HD, addressing gait detection challenges in chorea, while illustrating the broader utility of our approach in PD.

Given the challenges in collecting high-quality gait recordings from individuals with NDDs like HD, the scarcity of labeled data is a major barrier\cite{bansal2022systematic,alzubaidi2023survey}. Labeling data is time-consuming and expensive, often requiring expert clinicians to annotate recordings, which constrains dataset availability. Ethical considerations, participant burden, and logistical challenges also make collecting sufficient data difficult. This lack of comprehensive, labeled datasets restricts the development and validation of accurate models critical for identifying subtle deviations in gait patterns associated with movement disorders like HD. Overcoming this limitation is essential for advancing gait analysis research and improving outcomes for individuals affected by these diseases.

A promising solution to data scarcity is leveraging foundation models combined with transfer learning. This involves training a model on a surrogate task and adapting it to the task of interest through fine-tuning\cite{thrun1998lifelong}. Foundation models are becoming a major pillar of artificial intelligence\cite{devlin2018bert,liu2019roberta,ramesh2022hierarchical,openai2023chatgpt,changpinyo2021conceptual}. Typically, these models use self-supervised learning (SSL) to acquire rich feature representations that are fine-tuned for specific tasks using relatively small labeled datasets. This approach has shown promise in human activity recognition (HAR) and step detection, yielding superior performance over traditional methods\cite{small2023development,Yuan_2024,logacjov2024selfpab}. However, none of these studies have addressed the unique challenges of detecting gait abnormalities in HD and similar NDDs.

In this study, we show that existing methods for gait detection from wrist-worn accelerometers often incorrectly detect gait bouts of people with movement disorders such as HD. To mitigate this problem, we develop a novel approach for gait detection specifically designed for individuals with HD chorea in daily-living environments. Our approach leverages a pre-trained foundation model as the backbone for an advanced segmentation architecture, drawing inspiration from successful methodologies in computer vision\cite{chen2021transunet,cao2022swin}. We describe the model, show its enhanced gait detection abilities in HD (e.g., increasing the AUC from 35\% to 88\% for individuals with severe chorea), and demonstrate its generalizability by applying it to PD.

\section{Methods}\label{Methods}

\subsection{Dataset description}
Participants with HD and age-matched healthy controls (HC) were recruited as part of a study designed to evaluate the feasibility, acceptability, and utility of wearable activity monitors in people with early-mid stage HD. Tri-axial accelerometer signals were collected using wrist-worn devices during an in-clinic assessment. Participants completed a set of standardized laboratory-based functional activities, reflecting independent activities of daily living. The in-clinic assessment lasted approximately 30 minutes per participant. After this session, participants wore the GENEActiv (Kimbolton, UK) device on their wrist for the next 7 days in their daily environment to assess the performance of our model in daily-living. Details on device specifications, assessment protocols, and labeling procedures for the accelerometer signals are provided in the Supplementary Methods (SM).

\subsection{Model architecture and training}
\subsubsection{Foundation Model: Multi-Task Self-Supervised Learning}
We adopted a multi-task SSL pre-trained model introduced by \textcite{Yuan_2024} as the foundation for HAR. This SSL model was trained on the unlabeled UK Biobank (UKB) dataset, which comprises approximately 700,000 person-days of free-living activity data from over 100,000 participants, spanning seven days of wear for each participant. Notably, this model demonstrated superior performance in HAR, compared to previous HAR SSL models\cite{Yuan_2024}, likely because it leveraged a significantly larger dataset encompassing diverse natural human activities. In this approach, the tri-axial accelerometer data from wrist-worn activity trackers were used to train a feature extractor (FE) based on the ResNet-V2 architecture\cite{he2016identity}. The input to the network consisted of 10 second length windows of the 3-axis accelerometer data resampled to a frequency of $30 Hz$. The FE outputs a learned representation vector of 1024 dimensions, capturing essential information for subsequent activity recognition tasks.

\subsubsection{Baseline Model Architecture and Training Procedure}
\begin{figure}[h]
    \centering
    % Subfigure 1
    \begin{subfigure}[b]{\linewidth}
        \centering
        \includegraphics[width=0.9\linewidth]{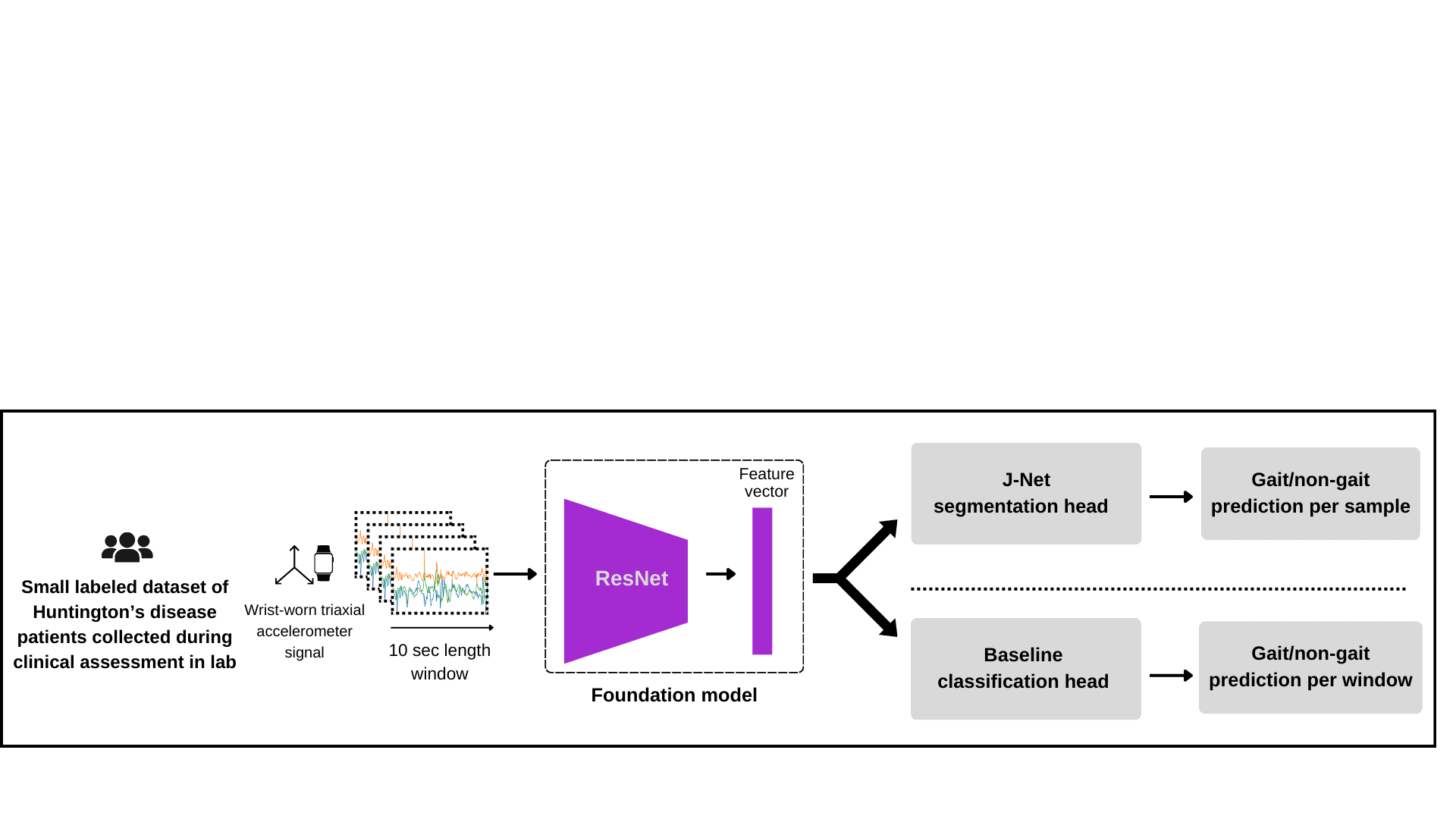}
        \caption{Illustration of gait detection algorithm for HD patients. The foundation model adopted from \textcite{Yuan_2024} based on ResNet and a feature extractor that are fine-tuned for gait detection of HD using the labeled data from the in-clinic assessment. Two options of task-specific heads are presented: a baseline classification head for predicting gait/non-gait per 10-second window, and a segmentation head for predicting gait/non-gait per sample within the 10-second window.}
        \label{Figure:fine-tuning procedure}
    \end{subfigure}
    
    \vspace{1em} % Adjust the vertical space between subfigures as needed
    
    % Subfigure 2
    \begin{subfigure}[b]{\linewidth}
        \centering
        \includegraphics[width=0.9\linewidth]{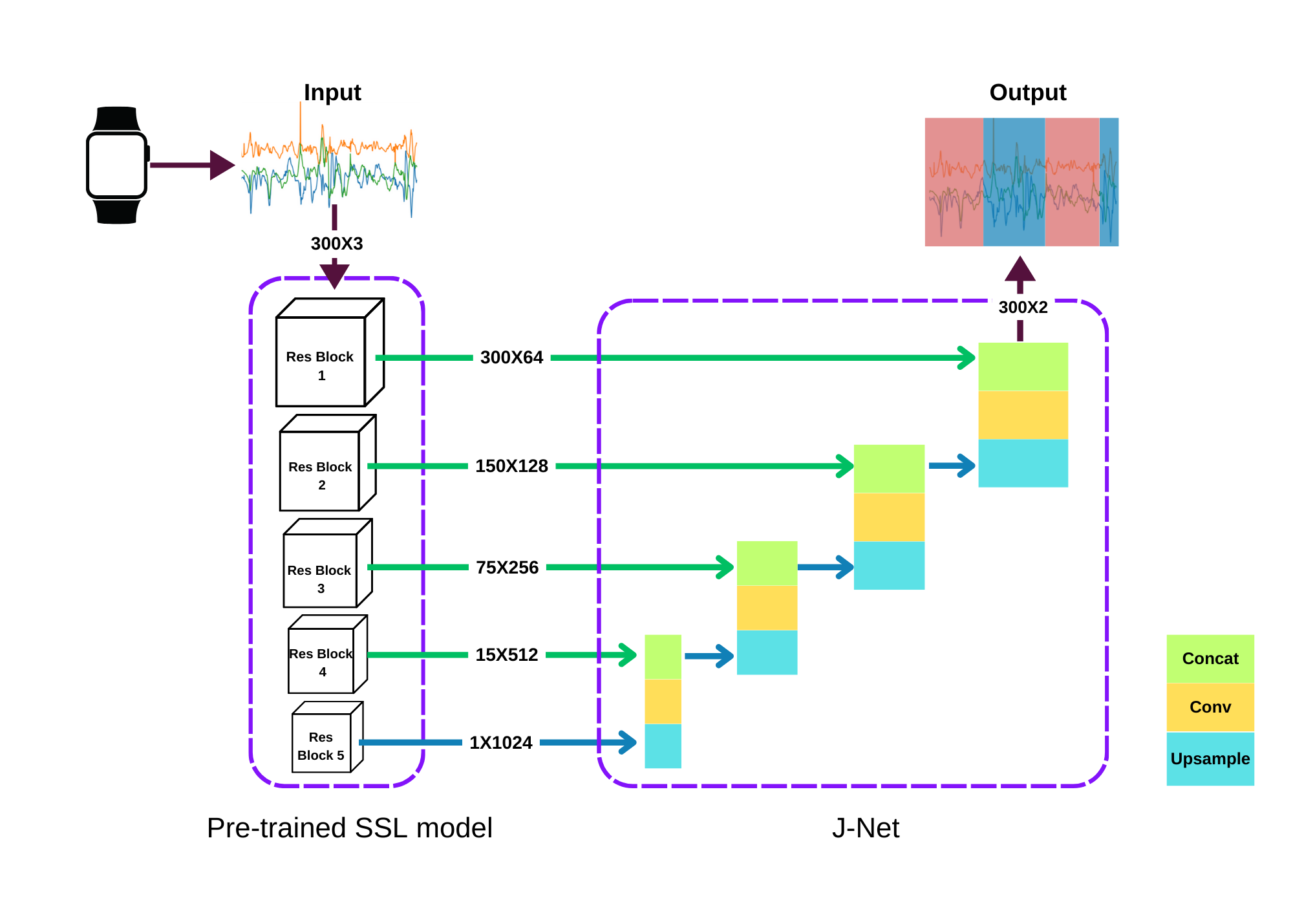}
        \caption{Illustration of the J-Net segmentation model architecture. Inspired by the U-Net architecture, the pre-trained feature extractor acts as the contracting path of U-Net, for down-sampling and enhancing feature channels. The expansive path incorporates up-sampling convolutions and concatenation of skip connections, followed by a convolution operation to map the feature vector to the desired classes. Input is a 10-second window, producing a segmented output distinguishing gait from non-gait activities.}
        \label{Figure:J-Net arch}
    \end{subfigure}
    
    \caption{Illustration of the gait detection algorithm for HD patients (top) and the J-Net segmentation model architecture (bottom).}
    \label{Figure:combined}
\end{figure}

To establish a baseline model, we adopted the architecture and training methodology originally outlined by \textcite{Yuan_2024} to fine-tune the model for downstream HAR tasks. We integrated the FE network with a binary gait detection head, consisting of a fully-connected (FC) layer and a sigmoid readout, facilitating the classification of each window as gait or non-gait. The pre-trained FE and the gait detection head were fine-tuned simultaneously using the HD dataset (See \cref{Figure:fine-tuning procedure}). The data preprocessing pipeline is described in the SM.

\subsubsection{J-Net Segmentation Model Architecture and Training Procedure}
To enhance the precision of gait detection based on wrist-worn accelerometer signals, we developed a novel segmentation head to combine with the pre-trained network architecture. This head modifies the network’s task to segmentation, effectively enabling sample-wise classification in contrast to the window-wise classification used in \textcite{Yuan_2024}. We developed the J-net architecture, shown in \cref{Figure:J-Net arch}, which is inspired by the U-net architecture\cite{ronneberger2015u} and previous similar implementations from the field of computer vision\cite{hatamizadeh2021swin,chen2021transunet}. In U-net, the data goes through a down sampling process followed by an up-sampling process. In J-net, the pre-trained foundation model replaces the down-sampling part of the U-net and then an up-sampling part is added to fine-tune the model for gait segmentation. Hence, the name J-net represents that the J-net is the right-hand side of the U-net. See the SM for details on the preprocessing steps, input format, and training procedure.

\subsection{Evaluation}
For the evaluation of our model trained on in-lab data, where ground-truth labels were available, we employed five-fold participant-wise cross-validation. The performance metrics were calculated based on this in-lab dataset. We computed a confusion matrix to analyze the gait detection results overall and stratified by chorea level. Additionally, we calculated the receiver operating characteristic area under the curve (ROC-AUC) to evaluate models of imbalanced data such as in gait detection. This assesses the performance of a classifier across all possible classification thresholds. For a detailed explanation of the AUC estimation method and the computation of confidence intervals, including the unbiased estimator and variance calculation and ablation study methods, please refer to SM.
We also calculated the recall and precision metrics to provide a nuanced understanding of model performance in gait detection across the different chorea levels. 

\subsection{Performance evaluation in Daily-living}
To evaluate the performance of J-Net in real-world scenarios, we used a dataset collected in daily-living. The gait detection models were applied to generate gait predictions. Since this data is unlabeled, we used gait predictions of the suggested segmentation model and the baseline classification model to perform a comparative analysis between the cohorts of HD and HC. We assessed hourly and daily walking percentages and walking time. Details on the statistical analysis, including the comparison between the HD and HC groups and the correlation with disease severity, are provided in SM.

\subsection{Performance evaluation in Parkinson's disease}
To evaluate generalizability to other NDDs, we used a dataset of 18 individuals with PD. These individuals wore a tri-axial accelerometer on their wrist for up to 10 days, with a sampling frequency of 25 Hz (Garmin, Olathe, KS, USA). A description of the dataset can be found in \cite{brand2022gait}. We compared the performance of the baseline classification model and J-Net on the PD dataset. Additional details on the evaluation with the PD dataset is in SM.

\section{Results}
In-clinic data collected from 25 participants with HD and 10 age-balanced HC participants were used for training and testing the models (Table S1).  Disease-specific status was quantified through the Unified Huntington’s Disease Rating Scale (UHDRS)\cite{kieburtz1996unified} including the TFC, Total Motor Score (TMS) and Functional Assessment (FA). The scores were administered by certified raters at each site  (Table S1). Data from daily-living of HD, HC and PD individuals were used to further evaluate and compare model performance. The dataset used in this study is publicly available and can be accessed at [https://doi.org/10.5281/zenodo.14384260]. Additionally, the code used is available on \href{https://github.com/DafnaSchwartz/hd_gait_detection_with_SSL}{GitHub}.

\subsection{Performance evaluation of baseline classification and J-Net segmentation models for gait detection}
The baseline classification model preprocessing produced 1,008 windows for the HD cohort and 766 for the HC cohort, while the segmentation model generated 5,920 windows for HD and 2,537 for HC, with each window containing 300 samples of 3-axis data. Examples of signals of HC, HD without chorea and HD with high chorea are illustrated in \cref{Figure:signals_exmp}. More details on the results of the preprocessing stage are provided in the Supplementary Results (SR).
Recall, precision, and ROC-AUC values for both the baseline classification model and the new J-Net segmentation model for HD and HC participants are summarized in \cref{tab:Combined_Comparison_with_CI}. When comparing the overall performance of the two models across chorea levels, J-Net exhibited significant improvements across all metrics. Specifically, there was a 10 percentage points increase in ROC-AUC for HD participants, reaching an AUC of $0.97 \pm 0.0006$, and a 4 percentage points increase in ROC-AUC for HC participants, reaching an AUC of $0.98 \pm 0.0004$. While both models performed better for HC, the segmentation model showed similar ROC-AUC values for HD and HC.
 \begin{figure}[h]
    \centering
    \includegraphics[width=\linewidth]{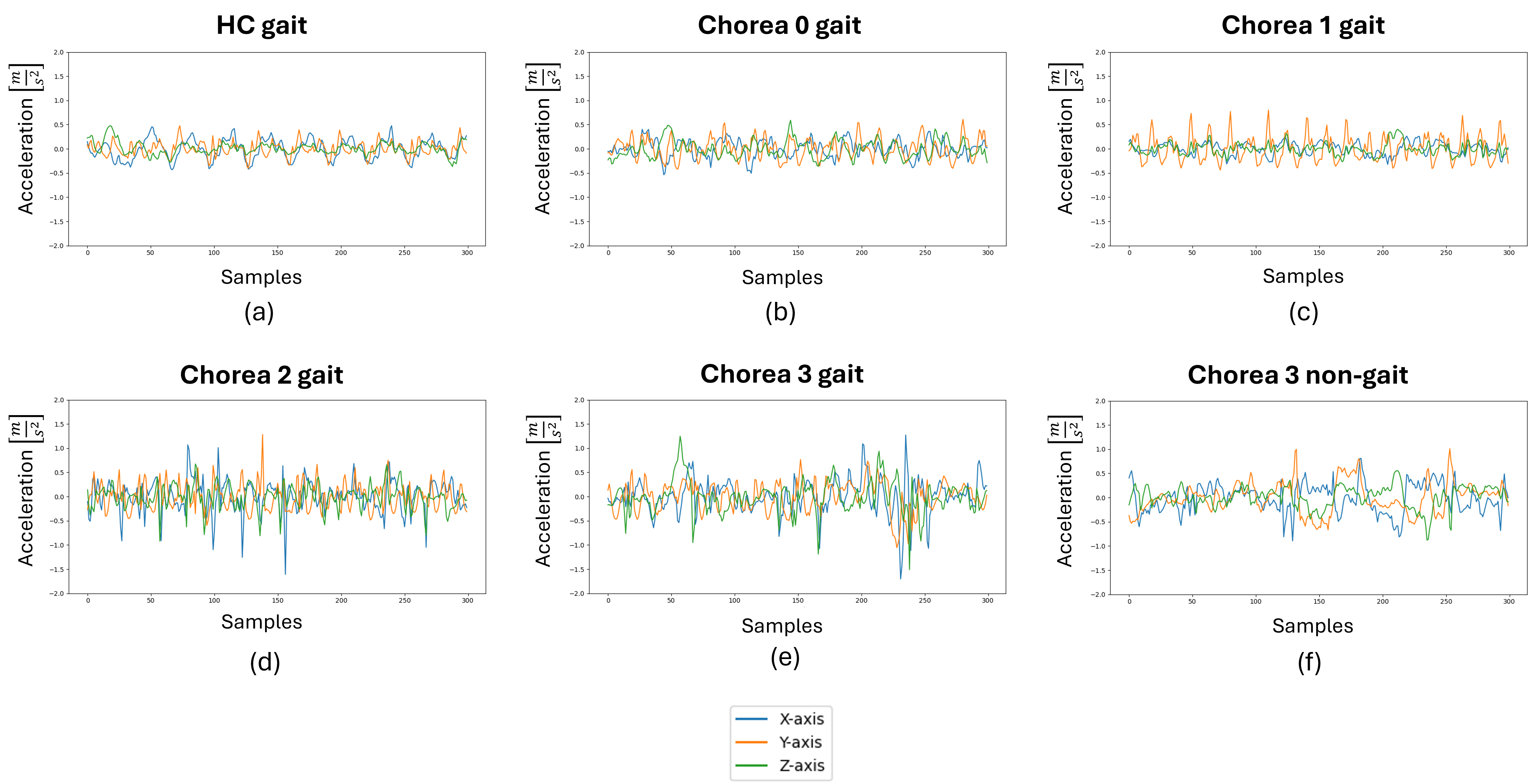}
    \caption{Examples of 10-second, 3-axis windows of accelerometer signals, with the Y-axis representing the vertical direction when the wearable device is properly worn on a vertically hanging hand. Shown are examples for: (a) Gait of a healthy control (HC), (b) Gait of an individual with Huntington's Disease (HD) without chorea (chorea 0), (c) Gait of an individual with slight chorea (chorea 1), (d) Gait of an individual with moderate chorea (chorea 2), (e) Gait of an individual with severe chorea (chorea 3), misclassified as non-gait by the baseline model but correctly classified as gait by the J-Net segmentation model, and (f) Non-gait (standing) of an individual with severe chorea (chorea 3), misclassified as gait by the baseline model but correctly identified as non-gait by J-Net.}
    \label{Figure:signals_exmp} 
\end{figure}

\begin{table}[htbp]
\centering
\caption{Comparison of model performances on in-lab data for HD and HC and AUC for different gait detection models across chorea levels. The J-Net segmentation model outperforms the baseline model across all metrics.}
\label{tab:Combined_Comparison_with_CI}
\begin{threeparttable}
\begin{tabular}{p{2.5cm} c c c}
\toprule
\textbf{Cohort/ Chorea Level} & \textbf{Metric} & \textbf{Baseline classification model} & \textbf{J-Net segmentation model} \\
\midrule
\multicolumn{4}{l}{\textbf{Performance over all chorea levels}} \\
\midrule
\multirow{3}{*}{HD} 
    & Recall    & 0.74 & 0.91 \\
    & Precision & 0.72 & 0.92 \\
    & ROC-AUC    & 0.87$\pm$0.02 & 0.97$\pm$0.0006 \\
\midrule
\multirow{3}{*}{HC}        
    & Recall    & 0.57 & 0.84 \\
    & Precision & 0.63 & 0.91 \\
    & ROC-AUC    & 0.94$\pm$0.04 & 0.98$\pm$0.0004 \\
\midrule
\multicolumn{4}{l}{\textbf{Performance per chorea level in the patients with HD\tnote{*}}} \\
\midrule
Level 0 (None)  & ROC-AUC & 0.94$\pm$0.05  & 0.97$\pm$0.001 \\
Level 1         & ROC-AUC & 0.84$\pm$0.07  & 0.98$\pm$0.002 \\
Level 2         & ROC-AUC & 0.80$\pm$0.08  & 0.96$\pm$0.002 \\
Level 3         & ROC-AUC & 0.89$\pm$0.06  & 0.95$\pm$0.003 \\
Level 4 (Severe)& ROC-AUC & 0.35$\pm$0.17  & 0.879$\pm$0.015 \\
\bottomrule
\end{tabular}
\begin{tablenotes}
    \item[*] Values are ROC-AUC$\pm$ 95\% confidence interval.
\end{tablenotes}
\end{threeparttable}
\end{table}

\cref{tab:Combined_Comparison_with_CI} summarizes the AUC and $95\%$  confidence intervals (CI) of the baseline model and the proposed segmentation model per chorea level. Fig. S1 displays a series of confusion matrices, organized in ascending order of chorea severity levels. As the chorea level increased, the ability of the models to accurately detect gait decreased. However, for the baseline model, the differences are larger between the chorea levels. Specifically, the reduction (degradation) in AUC between no chorea (chorea level 0) and highest chorea level (chorea level 4) with the baseline model was 58.5 percentage points. In contrast, with J-Net, the reduction was only 10 percentage points. In Chorea Level 4, the most severe stage marked by pronounced chorea, the baseline model demonstrated inadequate performance (AUC=0.35±0.17) while the J-Net still performed adequately (AUC=0.88±0.015). Ablation study results are in the SR.

%comapring the healthy and discussing generalizability 

\subsection{Daily living performance}

\begin{figure}[h]
    \centering
    \begin{tikzpicture}
        \node at (-1, 0) {
            \begin{subfigure}[b]{0.45\textwidth}
                \centering
                \includegraphics[width=\textwidth]{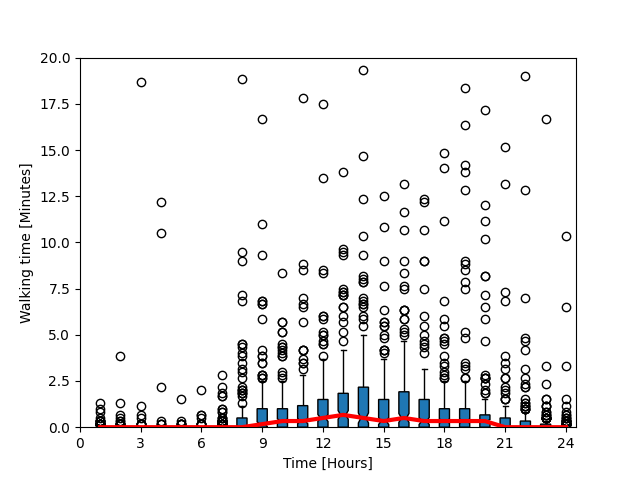}
                \caption{}
                \label{fig:minutes_walking_prob_per_hour_hc_class}
            \end{subfigure}
        };
        \node at (6.5, 0) {
            \begin{subfigure}[b]{0.45\textwidth}
                \centering
                \includegraphics[width=\textwidth]{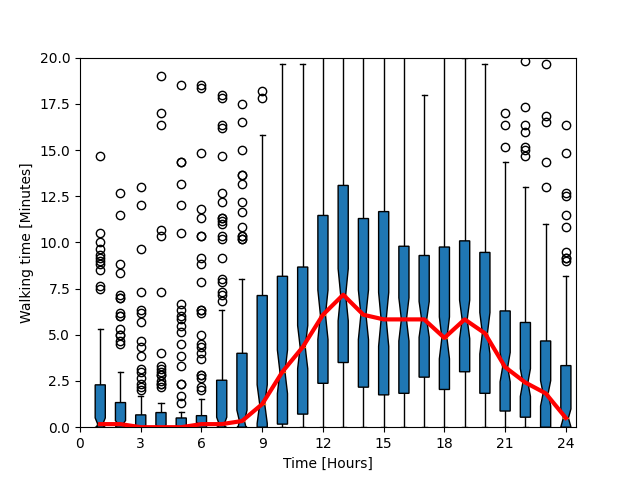}
                \caption{}
                \label{fig:minutes_walking_prob_per_hour_hd_class}
            \end{subfigure}
        };
        \node at (-1, -7) {
            \begin{subfigure}[b]{0.45\textwidth}
                \centering
                \includegraphics[width=\textwidth]{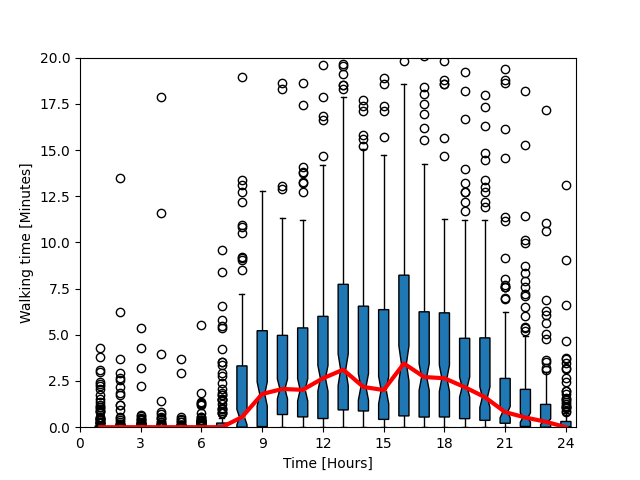}
                \caption{}
                \label{fig:minutes_walking_prob_per_hour_seg_hc}
            \end{subfigure}
        };
        \node at (6.5, -7) {
            \begin{subfigure}[b]{0.45\textwidth}
                \centering
                \includegraphics[width=\textwidth]{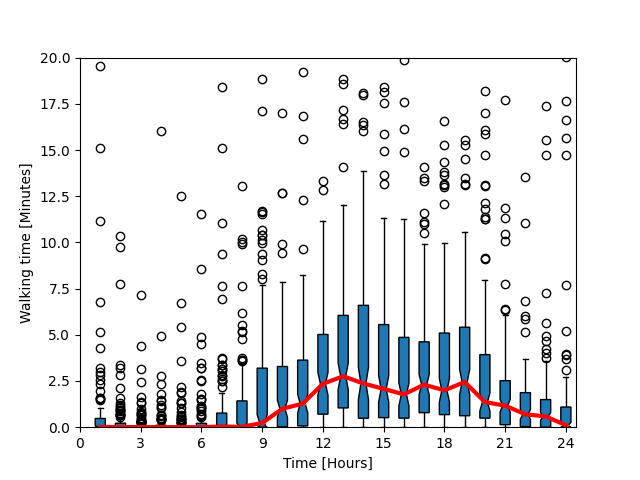}
                \caption{}
                \label{fig:minutes_walking_prob_per_hour_seg_hd}
            \end{subfigure}
        };

        % Adding labels
        \node[anchor=north] at (2.9, 3.4) {\textbf{Classification Model}};
        \node[anchor=north] at (2.9, -3.5) {\textbf{Segmentation Model}};
        \node[anchor=north] at (-1, 3) {\textbf{HC}};
        \node[anchor=north] at (6.5, 3) {\textbf{HD}};
        \node[anchor=north] at (-1, -4) {\textbf{HC}};
        \node[anchor=north] at (6.5, -4) {\textbf{HD}};

        % Adding Legend
        \node[anchor=north] at (2.9, -10) {%
               \begin{tabular}{m{1.5em} l}
                \raisebox{0.5ex}{\textcolor{red}{\rule{1.5em}{2pt}}} & Median walking time\\ 
            \end{tabular}
        };
    \end{tikzpicture}

    \caption{Diurnal walking patterns (minutes per hour). The top row shows classification model predictions for HC (a) and HD (b), while the bottom row shows segmentation model predictions for HC (c) and HD (d). As expected, Walking decreases at night. Counter-intuitively, the classification model indicates HD patients walked more than HC controls (median daily walking time, $p<0.005$), a finding not supported by the segmentation model ($p=0.234$).}
    \label{fig:distribution_over_24H}
\end{figure}

\cref{fig:distribution_over_24H} displays the diurnal pattern of walking time in minutes per hour as predicted by the two models for the HD and HC cohorts. The classification model under-estimated the walking time consistently throughout the day for HC and also counter intuitively identified more walking time in HD compared to HC (\cref{fig:minutes_walking_prob_per_hour_hc_class,fig:minutes_walking_prob_per_hour_hd_class})). This was not observed with the segmentation model (\cref{fig:minutes_walking_prob_per_hour_seg_hc,fig:minutes_walking_prob_per_hour_seg_hd}). Following the counter-intuitive result of the classification model for the hourly walking time, the median daily walking time per subject in the HD group was significantly greater than that of the HC group ($p<0.005$). Again, this counter-intuitive finding was not observed with the segmentation model ($p= 0.234$).  

\begin{figure}[h]
    \centering
    \includegraphics[width=0.6\linewidth]{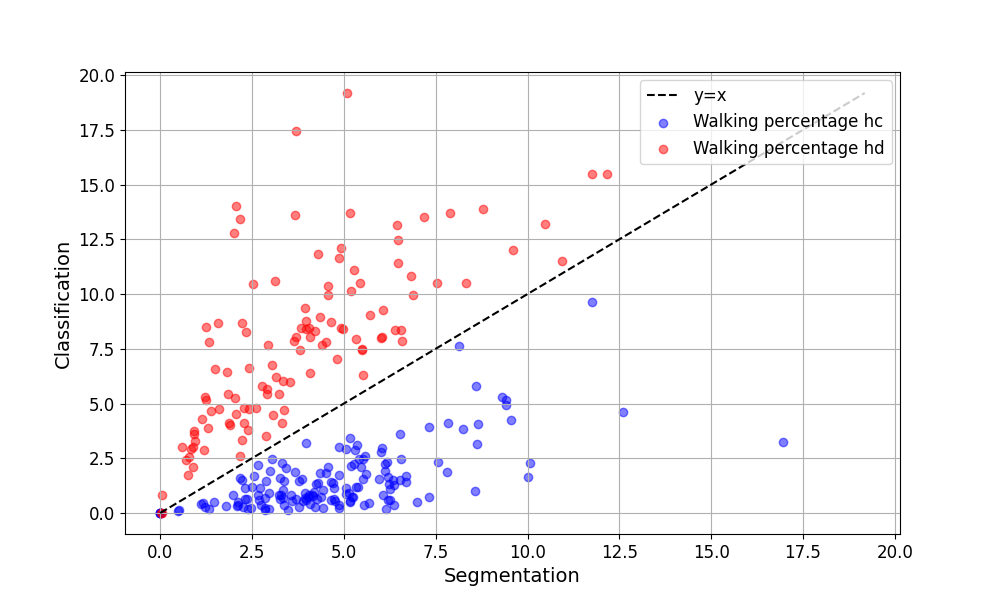}
    \caption{Daily walking percentage for each subject for each day predicted by the classification model and by the segmentation model. The y-axis denotes the classification predicted walking percentage, while the x-axis represents the segmentation predicted walking percentage. Each point corresponds to an individual on each day included in the analysis, with blue dots representing the HC group and red dots representing the HD group. The difference between the models leads to a clear separation between healthy controls (HC) and individuals with Huntington’s disease (HD). The dashed line represents the line y=x, where the predictions of the two models would be equal.}
    \label{Figure:compare_methods_HD_and_HC_scatter} 
\end{figure}

\cref{Figure:compare_methods_HD_and_HC_scatter} compares the daily walking percentage predicted by J-Net and the classification models for both HC and HD cohorts. The scatter plot shows a clear separation between the two groups. The classification model on the HD group tends to have higher values than the segmentation model, indicating an over-estimation of walking time for this group. However, for HC, the classification model has lower values than the segmentation model, indicating an under-estimation of walking time for this group. 

Fig. S4 illustrates the correlation between median walking time and a clinical measure of disease severity, the Unified Huntington's Disease Rating Scale – Total Motor Score (UHDRS-TMS). A negative correlation ($r = -0.52; p = 0.02$) was observed, indicating that individuals who walk more had lower (better) UHDRS-TMS scores, as expected.
\subsection{Performance in PD}
PD participants characteristics are summarized in Table S2. Fig. S5 shows the performance of our suggested J-Net model and the baseline classification model in detecting gait in PD in daily living settings. Both models showed improvement in all metrics after fine-tuning.  The fine-tuned J-Net model outperformed the baseline classification model in the AUC of the Precision-Recall (PR) curve, achieving 0.63 compared to 0.55. The AUC of the ROC curve was equal for both models (AUC-ROC = 0.92).
Additionally, J-Net fine-tuned on the PD dataset outperformed the best model from previously published work on gait detection in the same PD dataset\cite{brand2022gait}, which achieved an AUC of 0.89 for the ROC curve and 0.60 for the PR curve.

\section{Discussion}

In this study, we developed a novel deep learning model for gait detection using wrist-worn accelerometer signals that effectively addresses challenges in measuring gait for people with involuntary movements. By combining a pre-trained foundation model and an advanced segmentation architecture, our approach learns to detect gait even in the presence of irregular gait patterns seen in daily life, offering more reliable detection of gait bouts in individuals with movement disorders like HD.

Algorithms for HAR using wearable devices, particularly wrist-worn ones, are considered accurate and are integrated into popular smartwatch applications\cite{hammerla2016deep,bhat2020w,guan2017ensembles,lara2012survey}. Recent works describe robust and generalized algorithms for HAR in daily-living settings using pre-trained models fine-tuned for specific datasets \cite{small2023development,Yuan_2024,logacjov2024selfpab}.

Despite the robustness of these generalized algorithms in non-diseased populations, our study revealed that they do not generalize well to individuals with movement disorders, particularly HD and PD. When evaluating their performance on a dataset of individuals with HD, we observed a significant drop in accuracy as chorea severity increased. For the most severe chorea (level 4), the performance of the classification model was especially poor, with an AUC of just 0.35, highlighting its limited utility in detecting gait bouts at this severity level. This underscores the need for specialized models to handle the unique motor irregularities seen in severe cases.

To improve detection in severe chorea, we transitioned to the J-Net model, a segmentation-based architecture inspired by Swin-Net and TransUNet\cite{cao2022swin, chen2021transunet}. By combining a segmentation head with a pre-trained model, J-Net achieved significantly better performance across all chorea levels, with less degradation compared to the classification model. This ability to maintain higher performance even in the presence of severe chorea, demonstrates the model’s value and enables gait evaluation in daily-living settings and PD.

In daily-living evaluations, the segmentation model consistently outperformed the classification model in predicting walking time for both HD and HC cohorts. The classification model overestimated walking time in HD patients due to misclassifying chorea as gait and underestimated it for the HC cohort, leading to an incorrect suggestion that HD patients walked more than HC participants. In contrast, the J-Net model putatively provided more accurate estimates, with no significant differences in walking time between the cohorts. Additionally, the J-Net model's estimates of median walking time were negatively correlated with a clinical measure, indicating that individuals with less severe symptoms walked more. This suggests that the segmentation model can effectively differentiate true gait from movement artifacts like chorea, even in daily-living conditions.

Furthermore, when applied to PD data, J-Net once again outperformed the baseline classification model and also improved over previous published results\cite{brand2022gait}. This reinforces the potential of the J-Net architecture as a reliable tool for detecting gait in people with movement disorders across different neurodegenerative conditions. Fine-tuning the model to different datasets ensures its adaptability and effectiveness in various real-world contexts, particularly in daily-living settings where movement patterns are more complex and less structured than in clinical environments.

Our ablation studies provided insights into the strengths and limitations of J-Net and explored potential improvements (in the SR). The triple-window strategy for segmentation proved particularly effective, reducing inaccuracies at window boundaries and significantly enhancing event prediction accuracy, especially for detecting transitions between gait and non-gait activities. While our initial multi-task learning framework, which aimed to predict both gait and chorea levels, did not outperform the single-task model for gait detection, it offers a promising direction for future work: improving chorea detection through shared learning of gait and chorea features. 

Despite the promising results of our J-Net model, several limitations should be noted. One key limitation is the small size of the dataset used during the training phase. Standardizing data collection procedures for future studies will be essential to ensure high-quality video recordings, allowing more comprehensive labeling and training. Furthermore, the recruitment of participants with severe chorea levels (3 and 4) posed a challenge due to their lack of independence, resulting in fewer samples from this subgroup. Despite these limitations, the J-Net model showed significantly better performance than the baseline classification model in predicting gait, even for those with severe chorea. Finally, a challenge inherent in daily-living data is the validation of results. Future studies should aim to improve and expand the evaluation framework in daily-living environments, considering previous works that suggest solutions for this problem\cite{mazza2021technical,kluge2024real,mico2023assessing}.

In summary, the new J-Net architecture is a promising tool for gait detection in healthy adults and people with NDDs who experience involuntary movements using a single wrist-worn accelerometer. This approach outperformed existing methods by leveraging a novel combination of a foundation model and a segmentation network. For HD, we were able to accurately detect gait even in samples with severe levels of chorea, despite the noisiness of the signal and the scarcity of samples in this subgroup. In daily-living settings, a comparative analysis with baseline methods revealed that J-Net provides a more realistic estimation of daily walking time, which was also correlated with the clinical UHDRS-TMS score. These findings strengthen the validity and reliability of the model in real-world settings and pave the way for the objective assessment of daily-living gait in the presence of involuntary movements.

\section*{Acknowledgments} 
We thank Prof. Monica Busse at the Centre for Trials Research, Cardiff University in Cardiff, UK for assistance with data collection and invaluable help.

\clearpage
% Set double-spacing for the bibliography
\doublespacing

\printbibliography % Use biblatex command to print the bibliography

% Start Supplementary Sectio

\section{Supplementary Methods} \label{Supplementary_Methods}

\subsection{Dataset description}
Participants with HD and age-matched healthy controls (HC) were recruited from three clinical centers: (1) Teachers College, Columbia University, (2) the George Huntington Institute, M\"{u}nster, Germany, and (3) Cardiff University. The study was originally designed to evaluate the feasibility, acceptability, and utility of wearable activity monitors in people with early-mid stage HD. Tri-axial accelerometer signals were collected using wrist-worn devices (GENEActiv, Activinsights; 43 × 40 × 13 mm; weight: 16 g; 100 Hz sampling rate) in HD participants and HC during the in-clinic assessment. Participants completed a standardized laboratory-based circuit of functional activities, reflecting independent activities of daily living in the home. The overall in-clinic assessment lasted approximately 30 minutes for each subject. The accelerometer signals were recorded continuously during the assessment and, therefore, activities in between the standard tasks were also included in the signals. The in-clinic assessments were video-recorded, and these videos were subsequently used in the labeling procedure of the accelerometer signals. After the in-clinic session, the HD and HC participants were given a GENEActiv device to wear on their wrist for 24 hours per day over the next 7 days in their daily environment, with participants asked to continue their activities as usual. At the end of the 7-day period, participants removed the device and sent it back to the local clinical site. The data from the in-clinic assessment were used to train and validate the gait detection algorithm, and later, we used the daily-living dataset to assess the performance of our model in daily living.

\subsection{Labeling procedure}
The signals captured from the wearable sensor were synchronized with corresponding video recordings and were annotated by certified Unified Huntington’s Disease Rating Scale (UHDRS) motor raters. The start and end of activities including walking, sitting, standing and transitions between the activities (i.e., sitting down and standing up) were annotated. This labeling procedure resulted in continuous segments of the signal with an activity label. Similarly, we marked the start and end of every chorea episode together with the chorea level exhibited by the right arm (where the sensor was placed). Chorea levels ranged between 0 (none) to 4 (severe), reflecting the chorea severity, as defined by the UHDRS. Instances where the subject’s movements were unclear due to being obscured from the camera or positioned too far from it, resulting in uncertainty regarding the activity preformed or chorea rating, were designated as ’cannot be ranked’.

\subsubsection{Baseline Model Architecture and Training Procedure}
The data preprocessing pipeline for the baseline model included the following steps: 
\begin{enumerate}
    \item Applying a bandpass filter with a high cutoff frequency of $15 Hz$ and a low cutoff frequency of $0.2 Hz$
    \item Resampling the signal to  $30 Hz$
    \item Breaking the signals into 10-second windows
    \item Remove low-activity regions in the signal by calculating the standard deviation of the normalized energy within a window and discarding windows with an STD below $0.05$.
    \item Labeling each window a label of ”1” if more than $70\%$ of the window contained gait, and ”0” if more than $70\%$ contained non-gait.  Windows that did not meet these criteria were considered invalid and were removed.
\end{enumerate}  Consistent with \textcite{Yuan_2024}, we utilized the cross-entropy loss function for the task of gait detection. During training, we employed a batch size of 64 and utilized the Adam optimizer with a learning rate of \(1e^{-3}\) ~\cite{kingma2017adam}.

\subsubsection{J-Net Segmentation Model Architecture and Training Procedure}
The input to the network was a 10-second window (300 samples) and the output was a 10-second window segmented to gait and non-gait. In the preprocessing pipeline, we replicated the steps of filtering and resampling utilized in the baseline model. To address the challenge of accurately predicting events occurring at the edges of windows, we devised a strategy of creating batches of three consecutive windows with overlap of 5 seconds between the first window and the second window, and 5 seconds overlap between the second and third. In this way, we created an effective 10-second window that was padded with 5 seconds before and after. Each of the three windows was input to the feature extractor network and the outputs (including the output from the skip connections and the feature vector) were concatenated. The concatenated outputs were used as an input to the J-Net network that was then upsampled to the size of $3 \times 300$. The predictions of the network were for the middle window only. This triple window strategy allowed it to learn the context from 20 seconds overall (i.e.,10 seconds padded in 5 seconds at the beginning and end) and reduced potential inaccuracies near the window edges. Since this is a segmentation task, no majority rule was used, and the labels of each time sample were utilized. We employed the masked cross-entropy loss to handle samples labeled as ”-1”, indicating invalid data. We calculated the cross-entropy loss only of valid samples and normalized it by the number of valid samples. Consistent with the baseline model, we utilized a batch size of 64 and employed the Adam optimizer with a learning rate of \(1e^{-3}\)~\cite{kingma2017adam}.

\subsection{Evaluation}
We used the observation that the AUC is equivalent to the probability that a randomly chosen gait sample had a higher score than a non-gait sample (plus half of the probability of ties is added, if any), and that can be expressed as:
\begin{math}
    AUC = P[Y>X]+\frac{1}{2}P[Y=X]
\end{math} 
where X and Y are the scores of randomly chosen non-gait and gait samples, respectively. We used the unbiased estimator of the AUC presented in \textcite{cho2019confidence} and then the confidence interval was computed as a Wald-type interval based on the variance estimator of the ROC-AUC as suggested by \textcite{hanley1982meaning}.
We evaluated a model trained on healthy populations for detecting gait in individuals with HD. We assessed the performance of a baseline classification model fine-tuned with HC data by inferring on the HD dataset. Additionally, we evaluated the effectiveness of our segmentation technique and J-Net gait detection approach. We also compared the model's performance on HD patients versus HC subjects, providing insights into whether the model's adjustments resulted in specificity to HD or maintained generalizability across both healthy and affected populations, including those with chorea.

\subsection{Performance evaluation in Daily-living}
A comparison between the median walking time of the HD group and HC group was computed using a Student t-test. To further investigate the relationship between predicted walking time and disease severity, we calculated the median daily walking time for each subject. This metric was then correlated with the Unified Huntington’s Disease Rating Scale-Total Motor Score (UHDRS-TMS). We computed Spearman correlation coefficients to quantify the strength and direction of the relationship.

\subsection{Performance evaluation in Parkinson's disease}
An additional tri-axial accelerometer sensor was placed on the lower back. Labeling of the PD dataset into gait and non-gait segments was conducted based on “gold-standard” annotations from the lower-back sensor~\cite{galperin2019associations, galperin2020sensor}. We also fine-tuned the baseline classification model and our J-Net segmentation model with the PD dataset to assess their performance after adaptation to this specific cohort.

\section{Supplementary Results} \label{Supplementary_Results}
\renewcommand{\thefigure}{S\arabic{figure}}
\renewcommand{\thetable}{S\arabic{table}}
\setcounter{figure}{0} % restart figure counter
\setcounter{table}{0} % restart table counter
\subsection{Dataset}

\begin{table}[h]
\begin{threeparttable}[b]
\centering
\caption{Participant characteristics\tnote{*}}
\label{tab:Participant_characteristics}

\begin{tabular}{lccc}
\toprule
                                   & \textbf{HD participants} & \textbf{HC} & \textbf{P-value} \\
\midrule
\textbf{In-clinic}            & \multicolumn{3}{l}{}                     \\
\midrule
n                                  &       25          &        10      &   -      \\

\textbf{Demographics}              & \multicolumn{3}{c}{}                     \\
Age (yrs)                          &        56.7$\pm$11.2   &       58.6$\pm$10.3   &   0.64   \\
Gender (\% women)                  &         44$\%$        &            50$\%$  &      0.748\tnote{\textsuperscript{\dag}}  \\
Weight (kg)                        &          75.3$\pm$13.0       &        79.1$\pm$ 13.5 &   0.46      \\
Height (cm)                        &            173.6$\pm$9.2     &        170.8$\pm$10.6 &    0.47     \\

\textbf{HD clinical rating scores} & \multicolumn{3}{l}{}                     \\
UHDRS Total Motor Score (TMS)      &        41.1$\pm$16.5         &       -       &     -    \\
Total Functional Capacity (TFC)    &         10.6$\pm$2.1        &       -       &     -    \\
Functional Assessment              &         21.6$\pm$2.8        &        -      &     -    \\

\midrule
\textbf{Daily-living}         & \multicolumn{3}{l}{}                     \\
\midrule
n                                  &         21        &       28      &     -    \\

\textbf{Demographics}              & \multicolumn{3}{l}{}                     \\
Age (yrs)                          &        58.1$\pm$11.1         &       57.5$\pm$10.5     &     0.85    \\
Gender (\% women)                  &         50$\%$        &          54$\%$    &     0.8\tnote{\textsuperscript{\dag}}    \\
Weight (kg)                        &        72.8$\pm$13.7       &        80.7$\pm$14.2     &    0.06     \\
Height (cm)                        &          173.8$\pm$9.5       &            170.4$\pm$10.0  &    0.24     \\

\textbf{HD clinical rating scores} & \multicolumn{3}{l}{}                     \\
UHDRS Total Motor Score (TMS)      &        40.8$\pm$17.2         &        -      &    -     \\
Total Functional Capacity (TFC)    &         10.6$\pm$2.3        &        -     &     -   \\
Functional Assessment              &         21.6$\pm$3.7    &        -      &    -     \\

\bottomrule
\end{tabular}
\begin{tablenotes}
    \item [*] Values are mean$\pm$SD, except as indicated for gender. UHDRS, Unified Huntington Disease Rating Scale\cite{kieburtz1996unified}. UHDRS Total Motor Score (TMS) range: 0–124, with higher scores indicating greater motor impairment. Total Functional Capacity (TFC) range: 0–13, where higher scores indicate better functionality.
    \item [\dag] Chi-square test.
\end{tablenotes}

\end{threeparttable}
\end{table}
\cref{tab:Participant_characteristics} summarizes HD and HD characteristics. 
\subsection{Performance evaluation of baseline classification and J-Net segmentation models for gait detection}
 The preprocessing approach for the baseline classification model resulted in 1,008 windows of 3-axis data, each with 300 samples for the HD cohort and 766 windows for the HC cohort, utilized for training and evaluation. Meanwhile, the preprocessing approach for the proposed segmentation model generated 5,920 windows of 3-axis data, each with 300 samples for the HD cohort and 2537 windows for the HC cohort, utilized for training and testing. The differences between the number of windows originate in the approach to remove ambiguous windows (i.e., windows in which less than majority of $70\%$ from one activity) which we found to improve the classification model in a preliminary test. The preprocessing approach for the segmentation model allowed the inclusion of samples that were labeled as not valid in the classification training data.
 
\begin{figure}[h]
    \centering
    \includegraphics[width=0.35\linewidth]{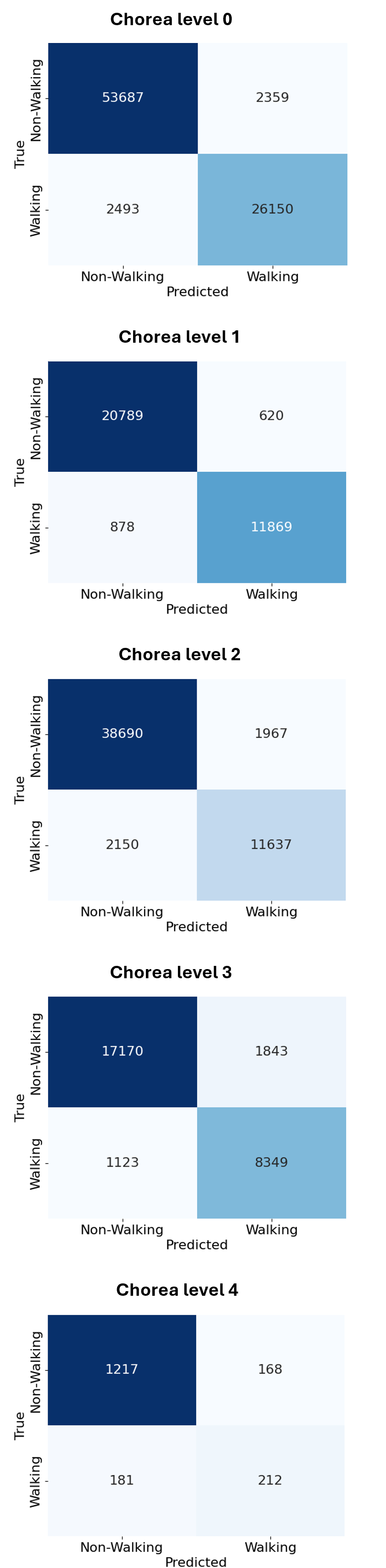}
    \caption{Confusion Matrix illustrating performance per chorea level of the J-Net segmentation model.}
    \label{Figure:confusion_matrix} 
\end{figure}

 \cref{Figure:confusion_matrix} displays a series of confusion matrices, organized in ascending order of chorea severity levels, from none to high. Each matrix corresponds to a specific chorea level and facilitates a comparative analysis of gait detection performance, stratified by the severity of chorea.

\subsection{Ablation studies}
\subsubsection{Evaluating the baseline classification model in HD using a model trained on a healthy population}
To evaluate the performance of the baseline model on detecting gait in HD without training on HD data, we tested the model that was fine-tuned on the HC data on HD data. The overall metrics across all chorea levels were lower except for precision. Recall was 0.41, precision was 0.79 and the ROCAUC was 0.75.
\subsubsection{Exploring segmentation methods for event prediction accuracy}
\begin{figure}[h]
    \centering
    \includegraphics[width=1\linewidth]{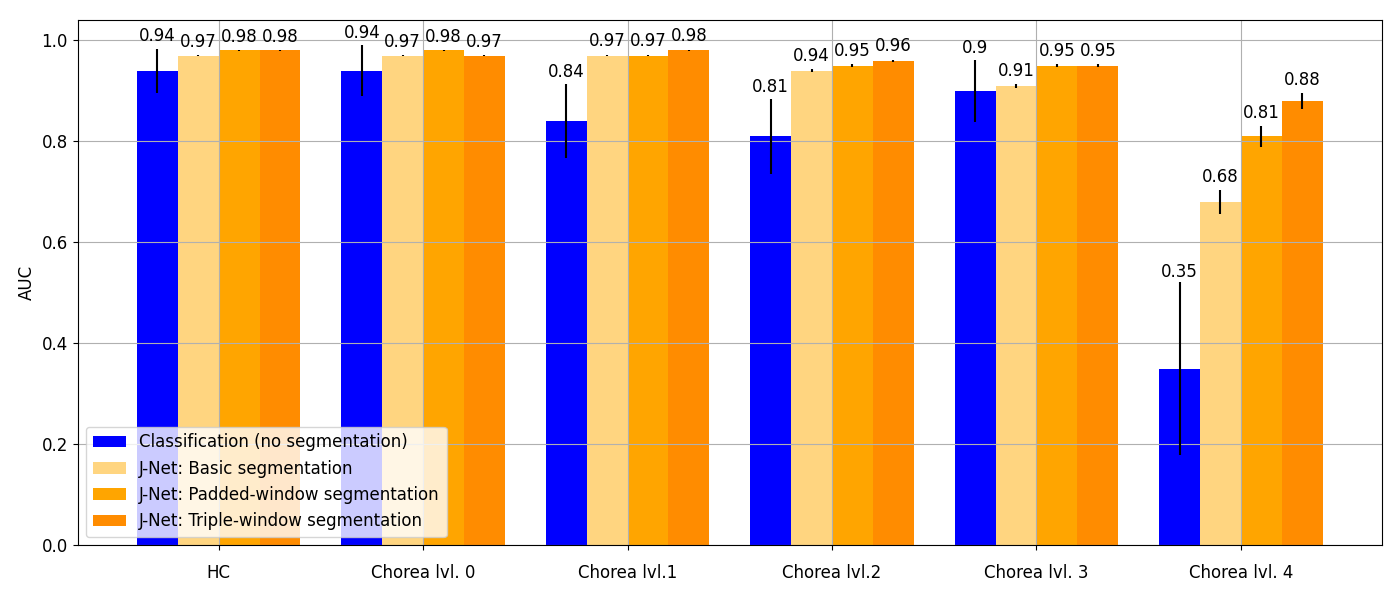}
    \caption{A bar graph comparing the Area Under the Curve (AUC) of three methods for enhancing event prediction accuracy at window edges within segmentation models. These methods include a basic segmentation approach, a padding and overlap method with effective 6-second windows, and a triple-window strategy. The latter, detailed in the Methods section, demonstrates superior performance in mitigating edge inaccuracies}
    \label{Figure:segmentation_methods_comparison} 
\end{figure}

In additional ablation analyses, we evaluated three distinct methods for enhancing the accuracy of event prediction at the edges of windows within segmentation models. Firstly, we examined a basic segmentation approach devoid of padding or overlap. Secondly, we introduced a method involving the creation of effective 6-second windows padded with 2 seconds of unclassified data at each end. This approach provided a buffer zone to mitigate potential inaccuracies near the window edges, along with a 4-second overlap between consecutive windows to ensure temporal continuity. Lastly, we employed a triple-window strategy where batches of three consecutive windows were utilized, incorporating overlaps to create an effective 10-second window. The details of this approach are described in the Methods section. \cref{Figure:segmentation_methods_comparison} presents a bar graph comparing the AUC of these methods alongside the classification baseline. We observed remarkable  improvements with the introduction of padding and overlap approach. Specifically, the triple-window approach, detailed in the methods section, emerged as the preferred choice due to its superior performance in addressing edge inaccuracies.

\subsubsection{Comparative analysis: Gait-Only versus multi-label approach}
In this ablation study, we transformed our proposed model into a multi-task framework, aiming to simultaneously segment gait/non-gait activities and determine the chorea level in each sample. This approach was compared to our previously established gait-only detection model. Consequently, each sample received both a gait/non-gait label and a chorea level label, enabling the model to learn concurrently for both tasks. By incorporating chorea level labels, we sought to enhance the model’s performance by leveraging shared and distinct features across tasks. Prior research has shown that multi-task learning can improve learning efficiency and prediction accuracy compared to training individual models separately \cite{baxter2000model,caruana1997multitask}. During the training phase, the overall loss function was a summation of losses from both tasks.

\begin{figure}[h]
    \centering
    \includegraphics[width=1\linewidth]{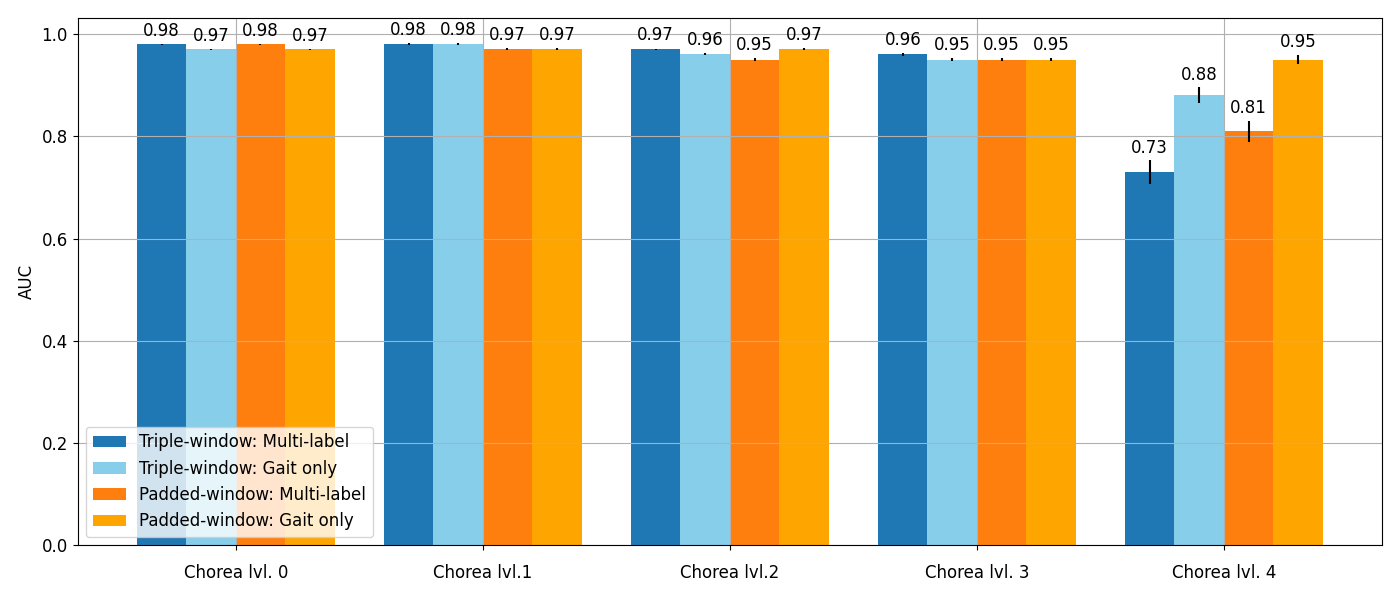}
    \caption{Comparison of gait-only and multi-label (gait and chorea) models within a multi-task framework for segmenting gait/non-gait activities and determining chorea levels. The performance evaluation, employing segmentation approaches (triple-window and padded-window), is depicted using a bar graph. Results indicate comparable high performance between the gait-only and multi-label models}
    \label{Figure:gait_only_vs_multi} 
\end{figure}
In evaluating the performance, we employed the segmentation approaches described earlier and compared the AUC of the gait-only and multi-label (gait and chorea) models that is presented in \cite{baxter2000model,caruana1997multitask}. Interestingly, the results showed high performance for both the gait-only and multi-label models, in both segmentation approaches (triple-window approach and padded-window approach), with no notable difference observed (\cref{Figure:gait_only_vs_multi}). This suggests that the multi-label approach holds promise for future studies, particularly in detecting chorea levels, as it serves as a solid foundation.

\subsection{Daily living performance}

\begin{figure}[h]
    \centering
    \includegraphics[width=0.6\linewidth]{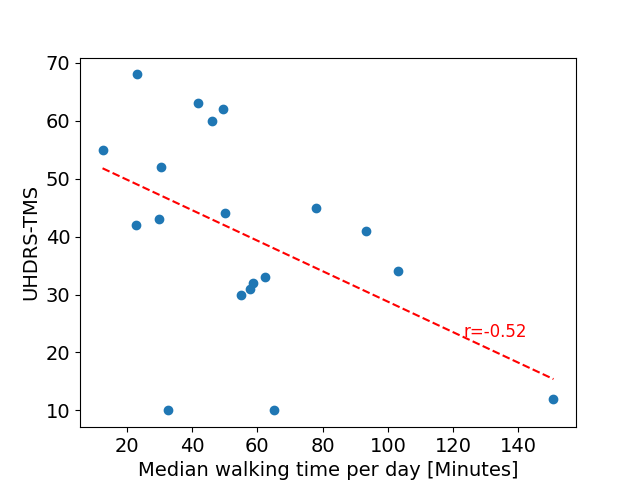}
    \caption{Relationship between walking time estimated by the J-net model and UHDRS-TMS}
    \label{Figure:UHDRS-TMS} 
\end{figure}

\subsection{Performance in Parkinson's disease}

\begin{table}[h]
\centering
\caption{Parkinson's disease participants characteristics}
\begin{threeparttable}
\begin{tabular}{lc}
\hline
\multicolumn{1}{|l|}{}                                  & \multicolumn{1}{c|}{\textbf{Parkinson's Disease (PD)}} \\ \hline
\multicolumn{1}{|l|}{n}                                 & \multicolumn{1}{c|}{18}                                \\ \hline
\multicolumn{1}{|l|}{\textbf{Demographics}}             & \multicolumn{1}{c|}{}                                  \\ \hline
\multicolumn{1}{|l|}{Age (yrs)}                          & \multicolumn{1}{c|}{72.7$\pm$8.4}                         \\ \hline
\multicolumn{1}{|l|}{Gender (\%women)}                   & \multicolumn{1}{c|}{44\%}                              \\ \hline
\multicolumn{1}{|l|}{Education (yrs)}                   & \multicolumn{1}{c|}{15.9$\pm$3.1}                              \\ \hline
\multicolumn{1}{|l|}{\textbf{PD clinical rating score}} & \multicolumn{1}{c|}{}                                  \\ \hline
\multicolumn{1}{|l|}{Hoehn and Yahr stage (1-4)}        & \multicolumn{1}{c|}{2.3$\pm$0.8}                        \\ \hline
\multicolumn{1}{|l|}{Disease duration (yrs)}         & \multicolumn{1}{c|}{5.5$\pm$4.05}                       \\ \hline
\multicolumn{1}{|l|}{Montreal Cognitive Assessment (0-30)}         & \multicolumn{1}{c|}{22.4$\pm$5.3}                       \\ \hline
\end{tabular}
\begin{tablenotes}
    \item [*] Values are mean$\pm$SD, except as indicated for gender characteristics. Hoehn and Yahr stage range: 1–5, with higher scores indicating more severe motor dysfunction. Montreal Cognitive Assessment (MoCA) range: 0–30, where higher scores indicate better cognitive performance.
\end{tablenotes}
\end{threeparttable}
\label{tab: Parkinson's disease participants characteristics}
\end{table}

\begin{figure}[ht]
    \centering
    \begin{subfigure}[b]{0.45\textwidth}
        \centering
        \includegraphics[width=\textwidth]{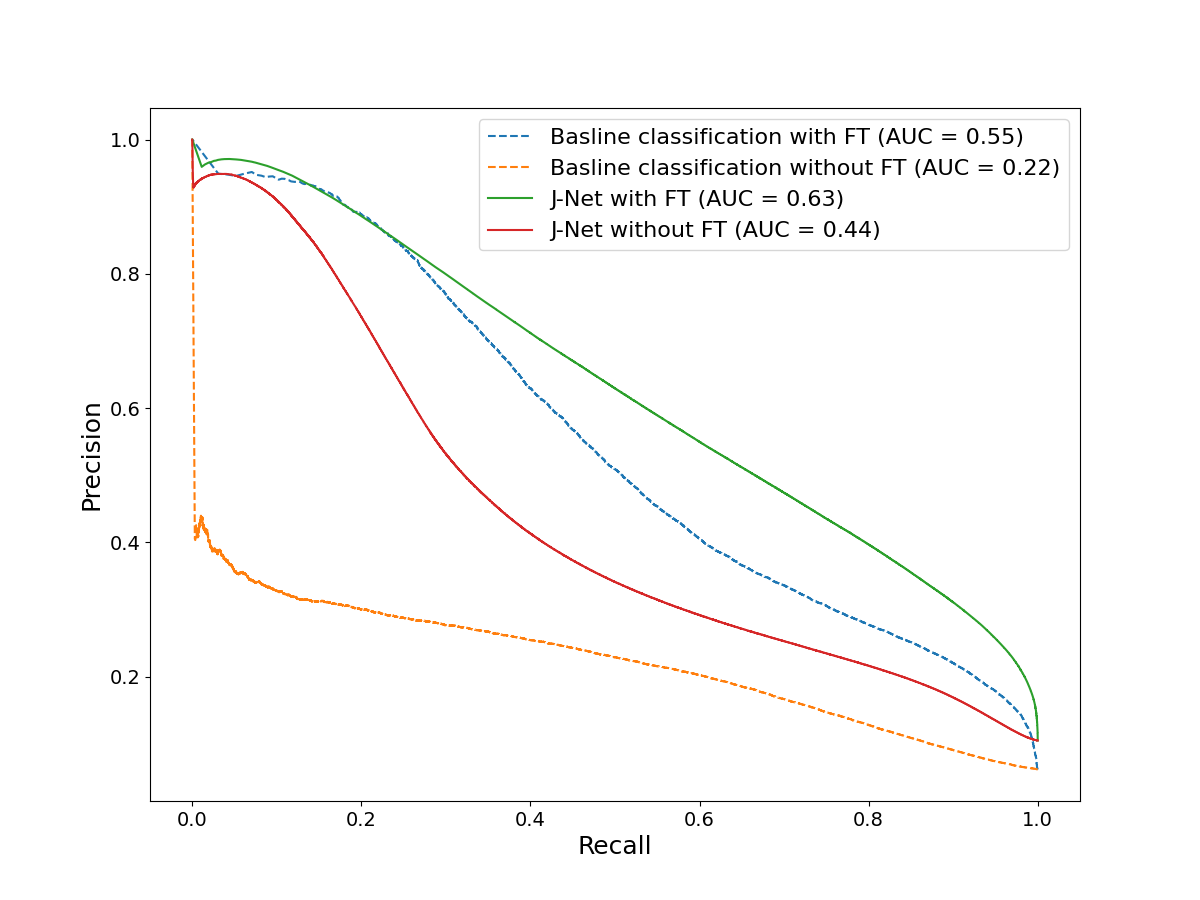}
        \caption{Precision-Recall Curve}
        \label{fig:precision_recall}
    \end{subfigure}
    \hfill
    \begin{subfigure}[b]{0.45\textwidth}
        \centering
        \includegraphics[width=\textwidth]{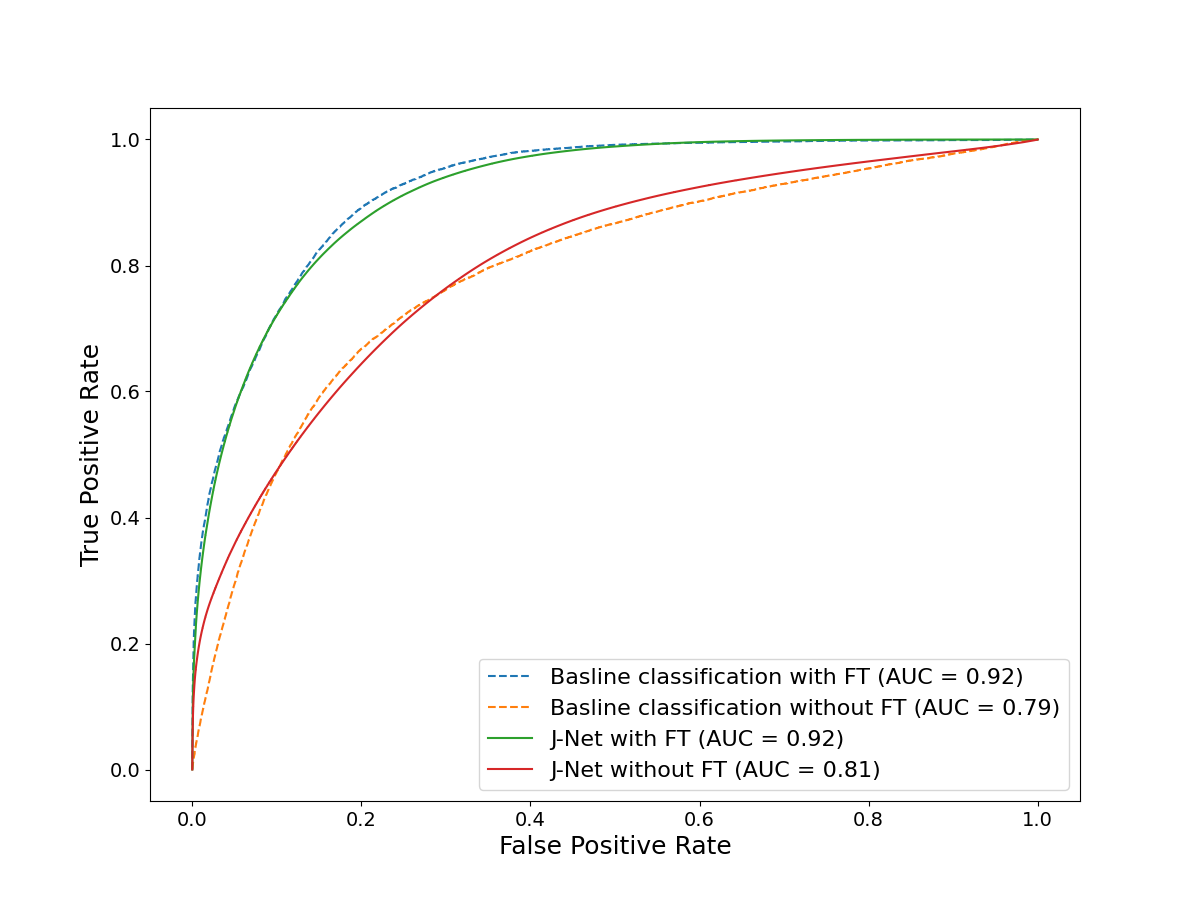}
        \caption{ROC Curve}
        \label{fig:roc}
    \end{subfigure}
    \caption{Performance of the models in PD. Comparing the baseline classification model and our J-Net segmentation model without fine-tuning (FT) and with FT with the PD dataset. AUC of the ROC curve and of the Precision-Recall (PR) curve for the J-Net outperformed also the best model in our previous work on the PD dataset \cite{brand2022gait}.}
    \label{fig:pd_performance}
\end{figure}

\end{document}